\documentclass[letterpaper, 10 pt, conference]{ieeeconf}  

\IEEEoverridecommandlockouts                              

\usepackage{cite}
\usepackage{amsmath,amssymb,amsfonts}
\usepackage{lipsum}  
\usepackage{algorithmic}
\usepackage{graphicx}
\usepackage{subcaption}
\usepackage{textcomp}
\usepackage{xcolor}
\usepackage{balance}
\usepackage[normalem]{ulem}

\newcommand{\fig}[1]{{\bf Fig. \ref{#1}}}

\usepackage{geometry}
\makeatletter
\let\NAT@parse\undefined
\makeatother
\usepackage[bookmarks=false]{hyperref}
\hypersetup{
    colorlinks   = true, 
    urlcolor     = blue, 
    linkcolor    = black, 
    citecolor    = blue, 
    pdftitle     = {Materials Matter: Investigating Functional Advantages of Bio-Inspired Materials via Simulated Robotic Hopping}, 
    pdfproducer  = {pdflatex}, 
}

\newcommand{\R}{\mathbb{R}}

\begin{document}

\title{\LARGE \bf
Materials Matter: Investigating Functional Advantages of Bio-Inspired Materials via Simulated Robotic Hopping
}
\author{Andrew K. Schulz$^{1,+}$, Ayah G. Ahmad$^{2,+}$, and Maegan Tucker$^{2}$
\thanks{+ These two authors contributed equally to this work.}
\thanks{$^{1}$Max Planck Institute for Intelligent Systems (MPI-IS), Stuttgart, Germany. email: {\tt\footnotesize aschulz@is.mpg.de}}%
\thanks{$^{2}$Georgia Institute of Technology, Atlanta, USA. emails: {\tt\footnotesize\{ayah, mtucker\}@gatech.edu}}}

\maketitle
\pagestyle{empty}

\begin{abstract}
In contrast with the diversity of materials found in nature, most robots are designed with some combination of aluminum, stainless steel, and 3D-printed filament. Additionally, robotic systems are typically assumed to follow basic rigid-body dynamics.
However, several examples in nature illustrate how changes in physical material properties yield functional advantages. 
In this paper, we explore how physical materials (non-rigid bodies) affect the functional performance of a hopping robot. In doing so, we address the practical question of how to model and simulate material properties. Through these simulations we demonstrate that material gradients in the leg of a single-limb hopper provide functional advantages compared to homogeneous designs. For example, when considering incline ramp hopping, a material gradient with increasing density provides a 35\% reduction in tracking error and a 23\% reduction in power consumption compared to homogeneous stainless steel. 

By providing bio-inspiration to the rigid limbs in a robotic system, we seek to show that future fabrication of robots should look to leverage the material anisotropies of moduli and density found in nature. This would allow for reduced vibrations in the system and would provide offsets of joint torques and vibrations while protecting their structural integrity against reduced fatigue and wear. This simulation system could inspire future intelligent material gradients of custom-fabricated robotic locomotive devices.

\end{abstract}

\section{Introduction} 
Research towards improving the functional capabilities of robots has traditionally focused on developing novel control methods\cite{raibert1984experiments, lewis2003robot, murray2017mathematical, westervelt2018feedback, spong2020robot, gehlhar2023review, chen2023bagging,srinivas2023busboy, adler2023teenager, o2024open}. For example, existing work has demonstrated the use of deep reinforcement learning for robotic walking \cite{peng2017deeploco, rudin2022learning, serifi2024vmp, margolis2024rapid}, optimization-based control for robotic-assisted locomotion on lower-limb prostheses and exoskeletons \cite{zhao2016multicontact, azimi2019model, horn2020nonholonomic, li2022natural, tucker2024synthesizing}, and large language models for dexterous manipulation \cite{ma2023eureka, li2024manipllm, jin2024robotgpt}. Along with the realization of new behavioral tasks, these novel control methods yield improved performance measures such as mechanical efficiency \cite{reher2020algorithmic} and reference tracking \cite{grandia2023doc}.

While developing such controllers is vital for improving robotic systems, these existing studies often neglect to analyze the effect of robotic \textit{design} on the overall performance. Moreover, while research exists towards optimizing kinematic design and mechanical structure to improve the performance of specific tasks \cite{maloisel2023optimal}, there is little research studying how material selection influences the associated functional capabilities. This discrepancy is magnified given the wide diversity of material properties observed in nature. Many biological examples demonstrate how various material properties improve functional capabilities\cite{elices2000structural,nepal_hierarchically_2023,schulz_elephant_2023,eder_biological_2018,barthelat_structure_2016,wegst_bioinspired_2015}, such as the material gradient of bone-tendon-muscle interfaces in the human body~\cite{li_review_2020,genin_functional_2009,badar_nonlinear_2025}. The material gradient properties have minimal stress concentration zones, providing less fatigue from cyclical tasks like walking~\cite{liu_functional_2017}. Additionally, bone, muscle, skin, and cartilage material properties vary significantly in nature~\cite{schulz_second_2025,schulz_skin_2022,wei_allometric_2017,schulz_elephants_2024}. In comparison, most robotic systems' limbs only consist of aluminum and stainless steel~\cite{ficht_bipedal_2021}. 
\begin{figure}[t]
    \centering
\includegraphics[width=0.48\textwidth]{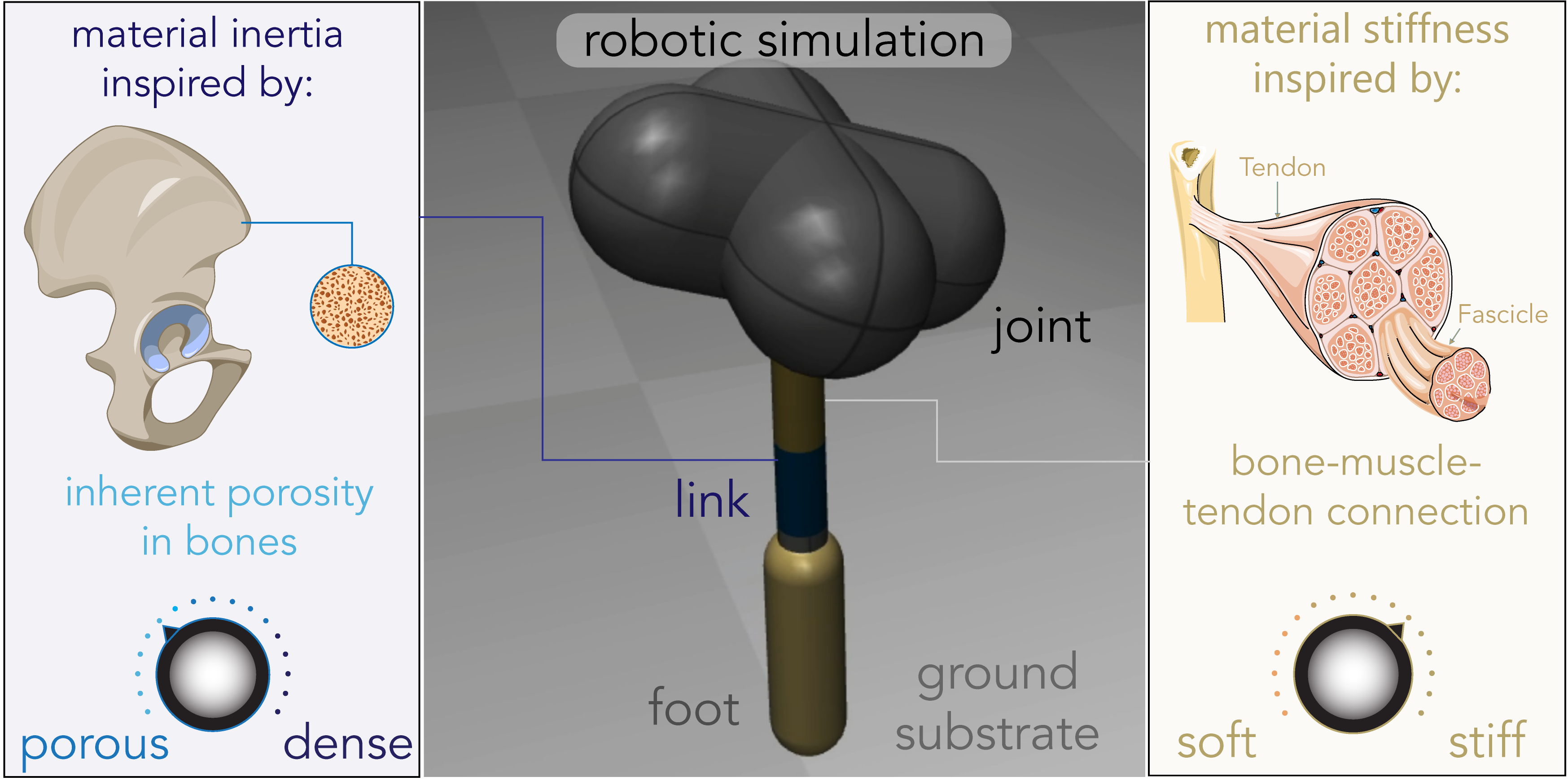}
    \caption{Our work explores how material properties (specifically porosities and moduli) influence the functional performance of a robotic system. The bio-inspiration behind this approach is the natural gradient of porosity and modulus across bone and bone-muscle-tendon connections.\vspace{-15pt}}
    \label{fig:1}
\end{figure}
Thus, to explore the space of material selection and its effect on robotic functional capabilities, we conduct a study that systematically evaluates the impact of material porosity and elastic modulus on the performance of a hopping robot. Explicitly, we take a bio-inspired approach and compare our results to known performance trade-offs of the bone-muscle-tendon interface in human bodies. 
This paper contributes:
\begin{enumerate}
    \item Fundamental background for bio-inspired materials as they relate to porosities and moduli in robotics (Section \ref{sec: materials}).
    \item A novel method for modeling materials in a robotic simulation suite (Section \ref{sec: methods}).
    \item Application of material simulation towards a one-legged robotic hopper (Section \ref{sec: hopper}).
    \item Analysis of functional changes resulting from shifts in material porosities and moduli demonstrated through a one-legged robotic hopper (Section \ref{sec: results}). 
\end{enumerate}

\begin{figure*}[t]
    \centering
\includegraphics[width=1\textwidth]{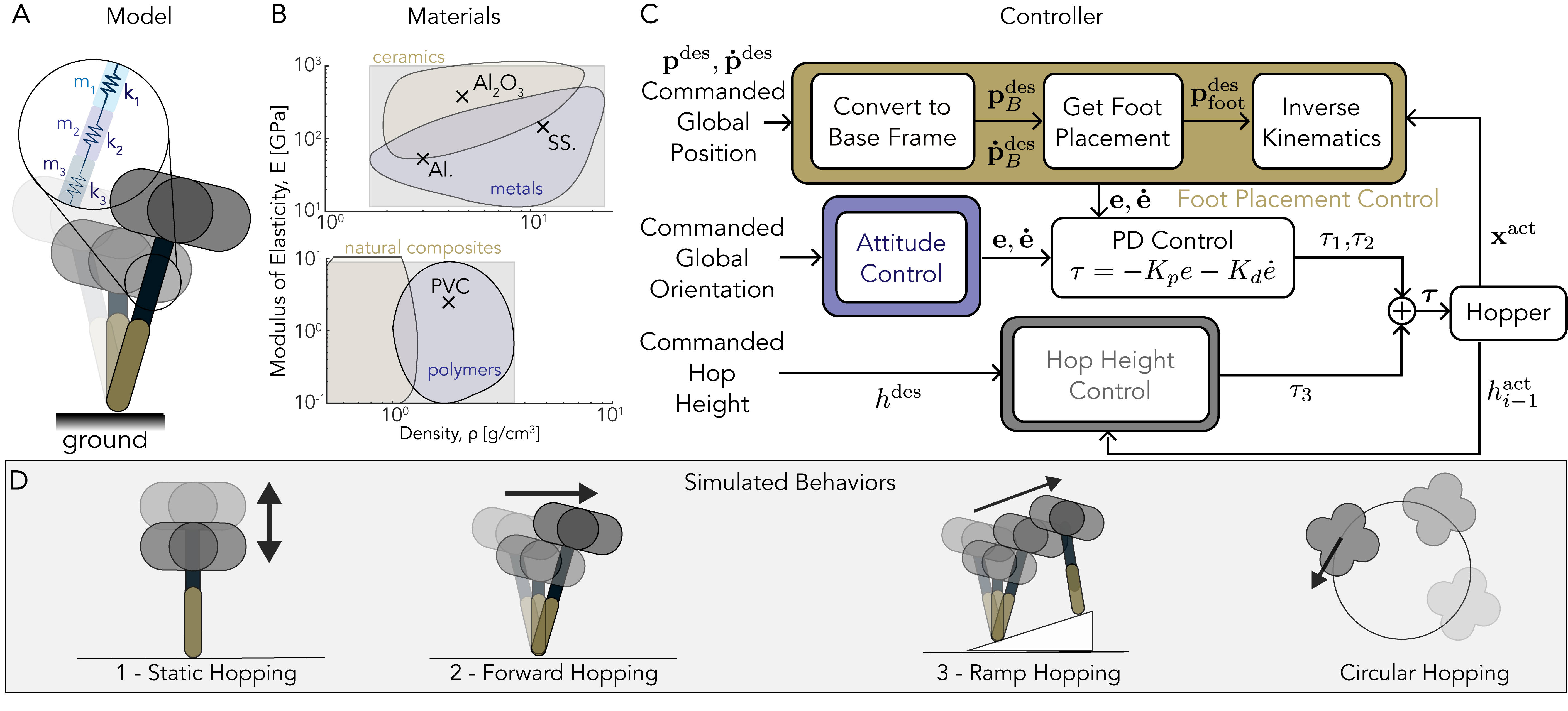}
    \caption{Framework for the experiments conducted with the robotic hopper shown in A) with the materials with mass $m$ and spring constant $k$ taken from modified Ashby plots shown in B) plotted of their density-modulus relationship. The robotic hopper moves via the controller shown in C) to perform the four simulated behaviors shown in D).\vspace{-14pt}}
    \label{fig:methods}
\end{figure*}

\section{Bio-inspiration of Biological Composites}
\label{sec: materials}
Many robotic systems are inspired by the natural geometries of humans~\cite{coradeschi_human-inspired_2006,ames_human-inspired_2014,abourachid_natural_2016} and animals~\cite{iida_biologically_2016, abourachid_natural_2016,li_terradynamics_2013,wang_mechanical_2023,gau_bridging_2023,phan_twist_2024}, mimicking their limb morphology to create joints linked as rigid-body systems. 
Despite significant advances in robotics control and practical applications in recent years~\cite{grau_robots_2021,abadia_neuromechanics_2025}, non-soft robotic simulation and construction communities often rely on assumptions such as the rigid-body assumption~\cite{nganga_accelerating_2021}. While inertia, friction, and geometric variables are essential~\cite{andrade_role_2021}, material quantities like porosity and modulus are ignored through the rigid-body assumption. Often, when material properties are used, they are assumed to be homogeneous or isotropic, and the material usage is limited to stainless steel~\cite{wang_development_2021} or aluminum~\cite{hussain_exoskeleton_2021,ficht_bipedal_2021}, which represent a small subset of the materials available in nature~\cite{wegst_mechanical_2004}. 

In biological systems, different proteins, sugars, and minerals come together to form anisotropic composites~\cite{eder_biological_2018,fratzl_natures_2007} where the form of the material connects to the function of the system~\cite{elices2000structural}. Composites like bones contain a porous core (\fig{fig:1}), which facilitates blood flow and shock absorption~\cite{morgan_bone_2018,siddique_lessons_2022}. Additionally, soft tissues such as muscles~\cite{kindig_effect_2015,lieber_structural_2004}, tendons~\cite{thorpe_chapter_2015,ping_mineralization_2022}, skin~\cite{graham_how_2019,boyle_morphology_2019}, and cartilage~\cite{kempson_patterns_1971} provide varying levels of stiffness (\fig{fig:1}) directly interfacing with bones to soften the forces and stresses transmitted to joints during locomotion~\cite{singh_mechanical_2021}. While robotic systems model these biological composites by their geometries~\cite{ames_human-inspired_2014}, their material complexities are often overlooked. 

Mechanical complexity and simulation simplicity are trade-offs, as are the material properties of modulus and porosity~\cite{jia_learning_2019}. The two most common materials in bi-pedal robots are stainless steel and aluminum ~\cite{lee_axiomatic_2006,ficht_bipedal_2021}, with aluminum's low density providing less mechanical stability, highlighting the trade-offs of material choice~\cite{luthin_framework_2021}. Recent innovations in bio-inspired materials interfaced with robotics have shown that because animals outperform robots in certain tasks, such as running \cite{burden_why_2024}, using things such as biomimetic muscles can improve the energy efficiency and adaptability of robotic systems \cite{buchner2024electrohydraulic}. 

\section{Methods of Simulating Stiff and Dense Material Linkages}
\label{sec: methods}
To study how material properties influence the functional performance of a robotic system, we develop a practical framework for modifying and simulating a continuum of material properties. This framework takes inspiration from the one-dimensional model of elasticity \cite{matsutani_eulers_2012}, to balance model complexity with simulation speed. 

This framework utilizes the simulation environment MuJoCo \cite{todorov_mujoco_2012}, but the same principles can apply to other environments like Gazebo \cite{koenig2004design} and Orbit \cite{mittal2023orbit}. 

\subsection{Simulating Material Porosity with Inertia}
In the current MuJoCo suite, the inertial properties of bodies are determined by the body's geometry and mass. If body mass ($m$) is not explicitly provided, it is computed using the formula $m = \rho V$, with $\rho$ denoting the uniform body density and $V$ denoting the body volume.

Material porosity can be related to density as the relationship between the overall bulk density and the particle density. Explicitly, material porosity is defined as $\Phi = 1 - \rho_b / \rho_s$ with $\rho_b$ denoting the overall bulk density, and $\rho_s$ denoting particle density. 
Material porosity can be connected to inertia by observing that for a fixed given volume $V_T$, the body mass scales with bulk density (i.e., $\rho_b = m/ V_{T}$). Thus, porosity scales with the mass of the link, which is directly proportional to inertia, i.e., $\Phi \propto I_{link}$. Lastly, to shift density across a link, it can be segmented into individual members as illustrated in (\fig{fig:methods}A).

\subsection{Simulating Material Moduli with Spring Stiffness}
While some packages exist for simulating flexible bodies, such as the elasticity plugin for MuJoCo \cite{mujoco_elasticity_plugin}, these packages are only computationally practical for very soft bodies such as elastomers and foams (E $\leq 100$ kPa, where E is the modulus of elasticity\cite{bhaskar_preliminaries_2023}). This limitation arises because they simulate the three-dimensional interaction forces between numerous particles, which becomes computationally expensive when considering stiff dynamics. The trade-off between simulation speed and material density/modulus is illustrated later in the manuscript.

To efficiently simulate material moduli, we embed several spring elements arranged in series within a link, allowing us to model mechanical fluctuation in the axial direction (\fig{fig:methods}A). This approach can be conceptualized as a one-dimensional elastica~\cite{matsutani_eulers_2012}. Additionally, this structure allows for adjusting the material density across the link.

To convert between a material's modulus of elasticity and the stiffness of each one-dimensional elastic element, we leverage the relationship $k = 3\text{E}I/L^3$ with $I$ denoting the beam's second moment of the area around the neutral axis. For cylindrical links, we use $I = \frac {\pi r}{4}$. Since the beam length and area remain constant throughout the models, $E \propto k$ for each of the members on the link. 

\begin{figure}[t]
    \centering
    \includegraphics[width=\linewidth]{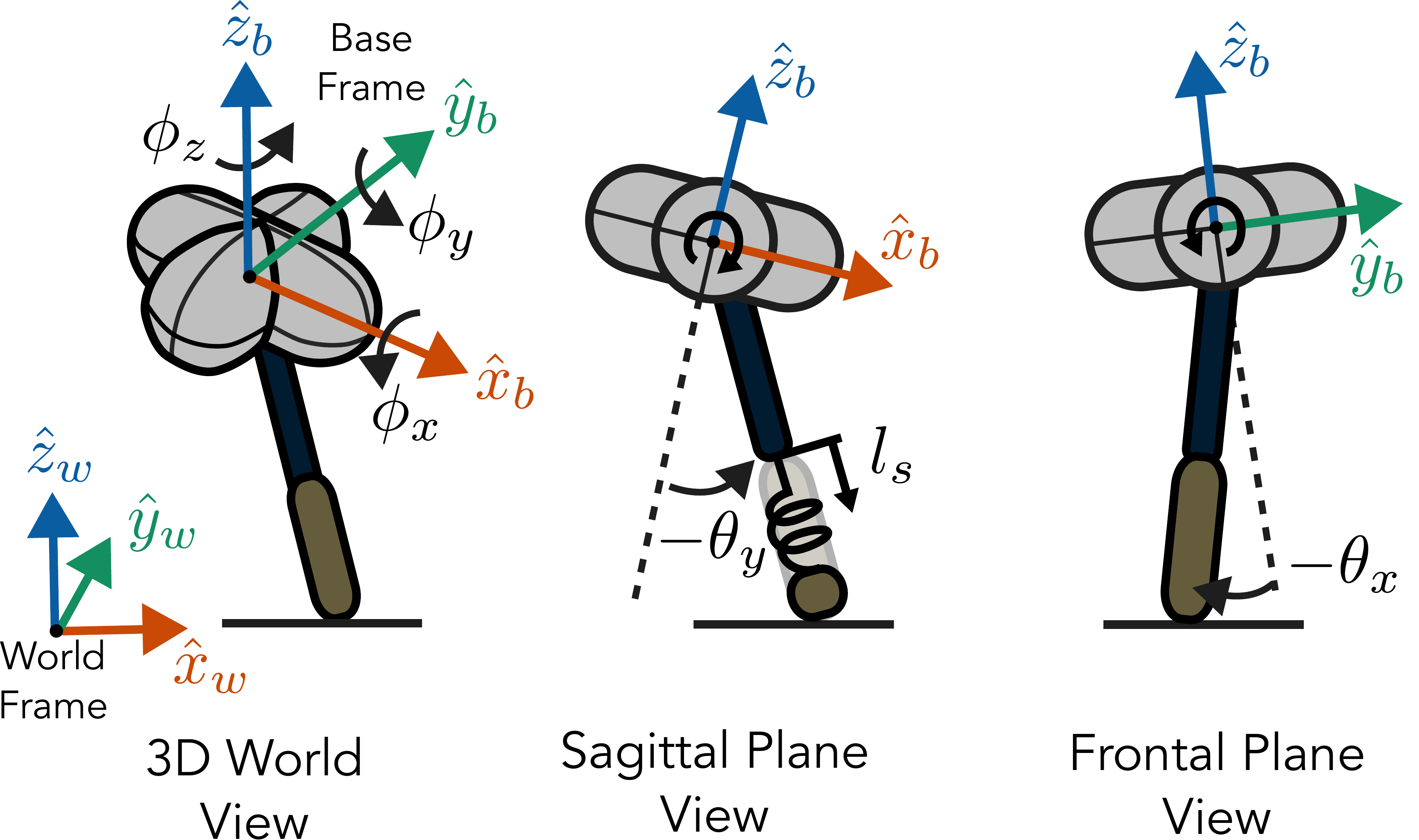}
    \caption{The configuration of the hopper is expressed as $\mathbf{q} = (\mathbf{p}, \boldsymbol{\phi}, \theta_x, \theta_y, l_s)$, in which $\mathbf{p}= (p_x, p_y, p_z) \in \R^3$ represents the position of the base frame relative to the world frame, $\boldsymbol{\phi} = (\phi_x, \phi_y, \phi_z) \in SO(3)$ denotes the orientation of the base frame relative to the world, $\{\theta_x, \theta_y\} \in \R$ represent the joint angles describing the angle of the leg about the axes $\hat{x}_b$ and $\hat{y}_b$, respectively, and $l_s \in \R$ denotes the extended length of the linear actuator controlling hop height.}
    \label{fig:configuration} \vspace{-14pt}
\end{figure}

\subsection{Bounds for Moduli and Density}
A standard tool used by material scientists to relate properties across materials is the Ashby plot\cite{wegst_mechanical_2004}. In general, Ashby plots display material properties on a log-log scale, with density ($\rho$) and modulus of elasticity (E) being the most common material properties compared. Materials with similar properties are often grouped together on the plot, as illustrated with metals and ceramics in \fig{fig:methods}B. 

Motivated by this method of comparing materials, we select the parameter bounds for material density and modulus using the material classes of metal-ceramics and polymer-natural composites. These material classes, as well as reasonable bounds, are illustrated in~\fig{fig:methods}B. 

\section{Simulating Material Properties for a Robotic Hopper}
\label{sec: hopper}
We demonstrate the material simulation approach towards the design and evaluation of a one-legged robotic hopper. Implementation steps are further documented in the paper's open-source repository\footnote{Code Repository: \href{https://github.com/dynamicmobility/mat-hopper}{https://github.com/dynamicmobility/mat-hopper}} and video \footnote{Video: \href{https://youtu.be/SEuEXRi80XU}{https://youtu.be/SEuEXRi80XU}}.

\subsection{Robot Configuration}

The hopping robot used in this paper is inspired by the one-legged hopping machine designed by Raibert et. al. in 1984~\cite{raibert1984experiments}. The configuration of the robot is illustrated in \fig{fig:configuration} and is denoted by $\mathbf{q} = (\mathbf{p}, \boldsymbol{\phi}, \theta_{x}, \theta_{y}, l_{s}) \in \mathcal{Q}$, and with the configuration derivative $\dot{\mathbf{q}} \in \mathsf{T} \mathcal{Q} = \mathcal{V} $, the full system state can be written as $\mathbf{x} = (\mathbf{q}, \dot{\mathbf{q}}) \in \mathcal{X} \triangleq \mathcal{Q} \times \mathcal{V}.$


\subsection{Controller}
The controller is based on the 3D hopping control framework proposed in \cite{raibert1984experiments}. As illustrated in \fig{fig:methods}C, the controller can be broken down into three main components: foot placement control, which is applied when the hopper is in the air (flight phase); attitude control, which is applied when the hopper is in contact with the ground (stance phase), and hop height control.

These separate control components together are responsible for obtaining the vector of actuation inputs  $\boldsymbol{\tau} = (\tau_1, \tau_2, \tau_3)^{\top} \in \mathcal{U}$. This vector of inputs corresponds to the actuation variables $\theta_x$, $\theta_y$ and $l_s$, respectively, with $\mathcal{U}$ being the admissible set of inputs (due to actuation constraints). 

\subsubsection{Foot Placement Controller}

When the hopper is in the flight phase, the desired foot position is computed using the law
\begin{align}
    \mathbf{p}_{\textrm{foot}}^{\textrm{des}} = \frac{T_{ST}}{2}(\dot{\mathbf{X}})+ K_1 (\dot{\mathbf{X}}-\dot{\mathbf{X}}^{\textrm{des}}),
    \label{eq: pdes}
\end{align}
with $\mathbf{p}_{\textrm{foot}}^{\textrm{des}} = [p_{\textrm{foot},x}^{\textrm{des}}, ~p_{\textrm{foot},y}^{\textrm{des}}]^{\top}$ denoting the desired position of the foot in the $x-$ and $y-$axes relative to the local body frame, and $\mathbf{\dot{X}} = [\dot{p}^{\textrm{act}}_x, \dot{p}^{\textrm{act}}_y]^{\top}$ and $\mathbf{\dot{X}}_d = [\dot{p}^{\textrm{des}}_x, \dot{p}^{\textrm{des}}_y]^{\top}$ denoting the actual and desired velocity, respectively, of the base in the $x$ and $y$ coordinate axes relative to the local base frame. Additionally, $T_{ST} \in \R_{+}$ denotes the duration of the stance phase and $K_1 \in \R_+$ denotes a constant gain.
\begin{figure}[t]
    \centering
\includegraphics[width=0.5\textwidth]{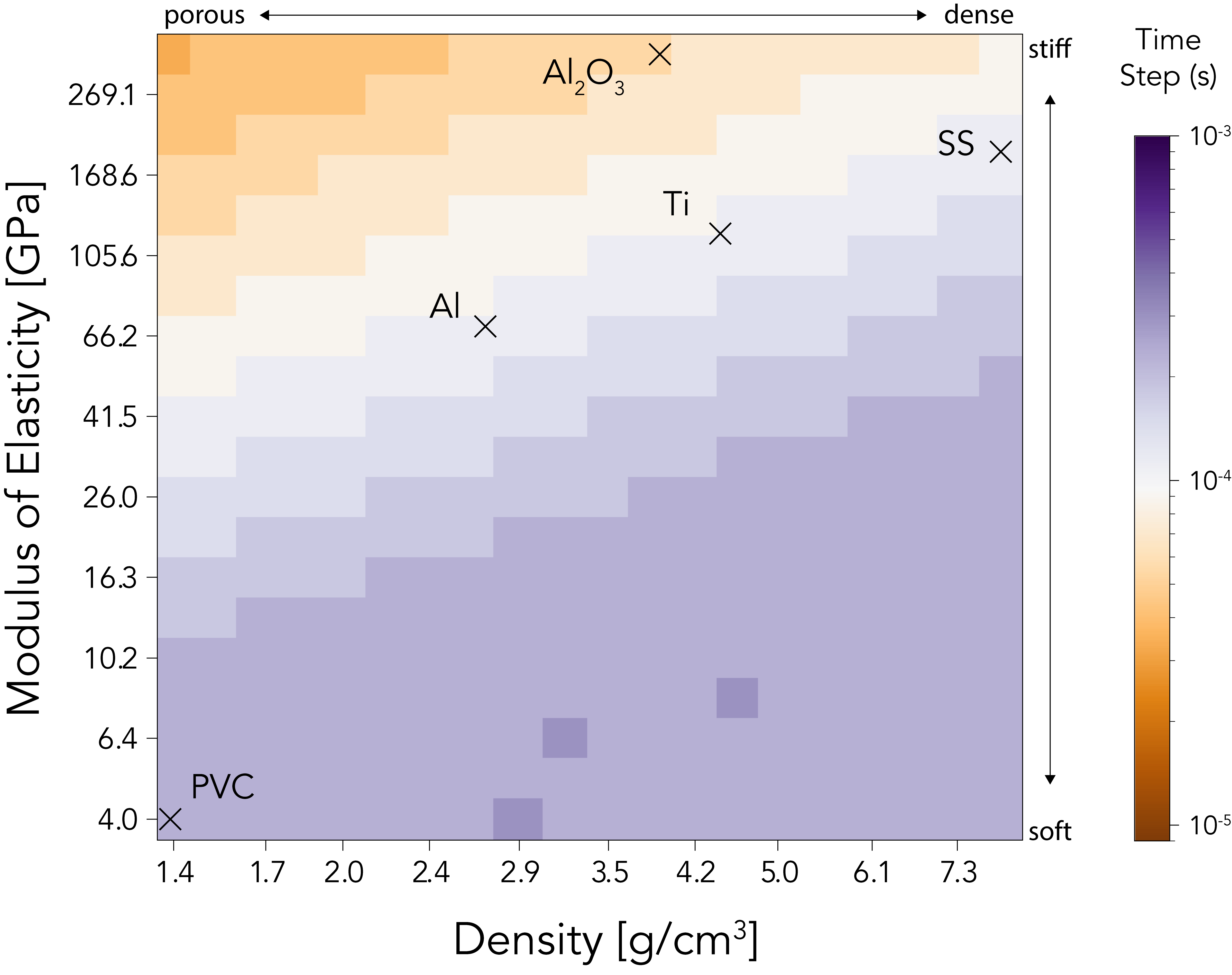}
    \caption{The heat map illustrates the effect of material modulus and density on the time step required to maintain stable forward integration of the dynamics. The results show that dense/soft materials allow for faster simulation speeds, while porous/stiff materials require more frequent computation of the dynamics, resulting in slower simulation speeds.\vspace{-14pt} }
    \label{fig:heatmap}
\end{figure}

In Equation \eqref{eq: pdes}, the desired velocity $\dot{\mathbf{X}}^{\textrm{des}}$ is obtained by
\begin{align}
    \dot{\mathbf{X}}^{\textrm{des}} &= \textrm{min} \{ -K_2(e(\mathbf{X})) + K_3 (\mathbf{\dot{X}}^{\textrm{des}}), \mathbf{\dot{X}}_{\textrm{max}} \},
\end{align}
where $e(\cdot) := (\cdot)^{\textrm{act}} -  (\cdot)^{\textrm{des}}$ denotes the error between the actual and desired values of the argument, $K_2$ and $K_3$ denote constant gains and $\mathbf{X}_{\textrm{max}}$ is a constant that restricts the maximum constant velocity for the robot. 

After computing $\mathbf{p}_{\textrm{foot}}^{\textrm{des}}$, the joint positions ($\theta_x$ and $\theta_y$) required to achieve this desired foot placement are calculated using inverse kinematics. The computed desired joint angles are denoted as $\theta_x^{\textrm{des}}$ and $\theta_y^{\textrm{des}}$
These desired joint positions are enforced during the flight phase using PD control~\cite{aastrom2021feedback}
\begin{align}
    \tau_1 &= -K^f_p(e(\theta_x)) - K^f_d(e(\dot{\theta}_x)), \\
    \tau_2 &= -K^f_p(e(\theta_y)) - K^f_d(e(\dot{\theta}_y)),
\end{align} 
where $\{K^f_p, K^f_d\} \in \R_{+}$ are constant gains.

\subsubsection{Attitude Control}
The attitude of the hopper base frame can only be controlled when the hopper is in the stance phase. The following control law is applied to the rotational joints
\begin{align}
    \tau_1 &= K^s_p(e(\phi_x)) + K^s_d(e(\dot{\phi}_x)), \\
    \tau_2 &= K^s_p(e(\phi_y)) + K^s_d(e(\dot{\phi}_y)).
\end{align}
Here, $\{K^s_p, K^s_d\} \in \R_+$ are proportional and derivative gains specifically tuned for the stance phase, distinct from the gains used during the flight phase.

\subsubsection{Hop Height Control}
The linear actuator connected to the physical spring is commanded to apply an impulsive force after the leg makes contact with the ground. As shown in \cite{raibert1984experiments}, a given constant impulsive force will result in a constant hop height of the overall system. An adaptive control law is used to enforce a constant hop height of one meter across all materials. The impulsive force applied to the linear actuator controlling $l_s$ is defined to be
\begin{align}
    \tau_3 = 
    \begin{cases}
        F_{i} = F_{i-1} + K e_{i-1}, & \textrm{if } F_z > 0, ~\dot{p}_z > 0\\
        0, & \textrm{otherwise}.
    \end{cases}
\end{align}
Here $e_{i-1} \in \R_+ := h^{\textrm{des}} - h^{\textrm{act}}_{i-1}$ denotes the error between the desired hop height and the actual hop height of the last hop cycle, $F_{i}$ denotes the constant linear force applied during hop cycle $i \in \mathbb{Z}$. In other words, the linear force is adaptively updated to minimize the error between the desired hop height, $h^{\textrm{des}} = 1$ and  the hop height of the previous hop cycle, $h^{\textrm{act}}_{i-1} \in \R_+$. This linear force is only applied when the hopper leaves the ground, mathematically captured by the conditions $F_z > 0$ (i.e., the vertical ground reaction force is positive) and $\dot{p}_z > 0$ (i.e., the vertical velocity is positive).

 \begin{figure}[t]
    \centering
     \includegraphics[width=0.98\linewidth]{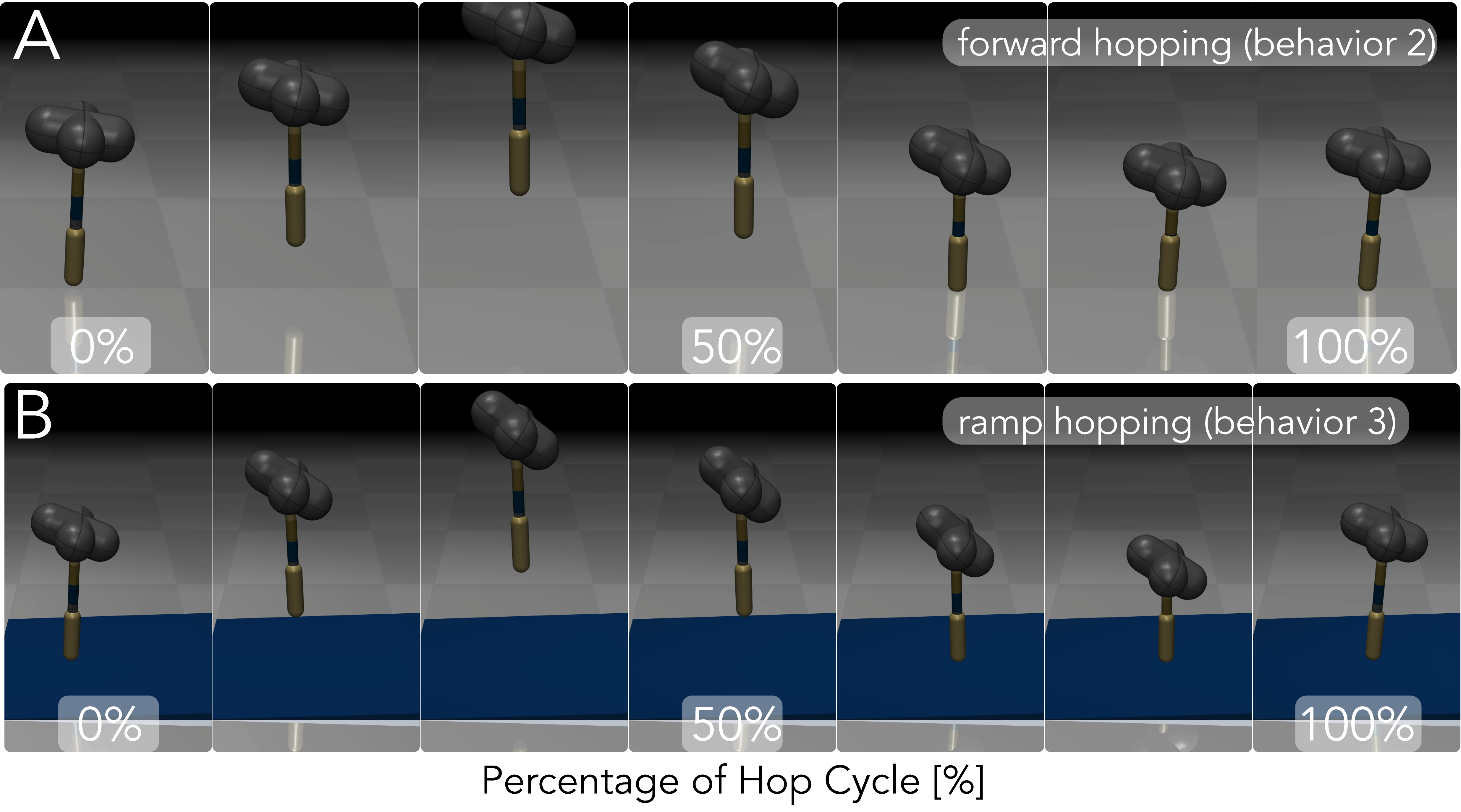} 
    \caption{Gait tiles illustrating the hopping motion for A) forward hopping and B) up-ramp hopping on 3.5\% grade.\vspace{-14pt}}
    \label{fig:tiles}
\end{figure}

\begin{figure*}[t] 
    \centering
\includegraphics[width=1\textwidth]{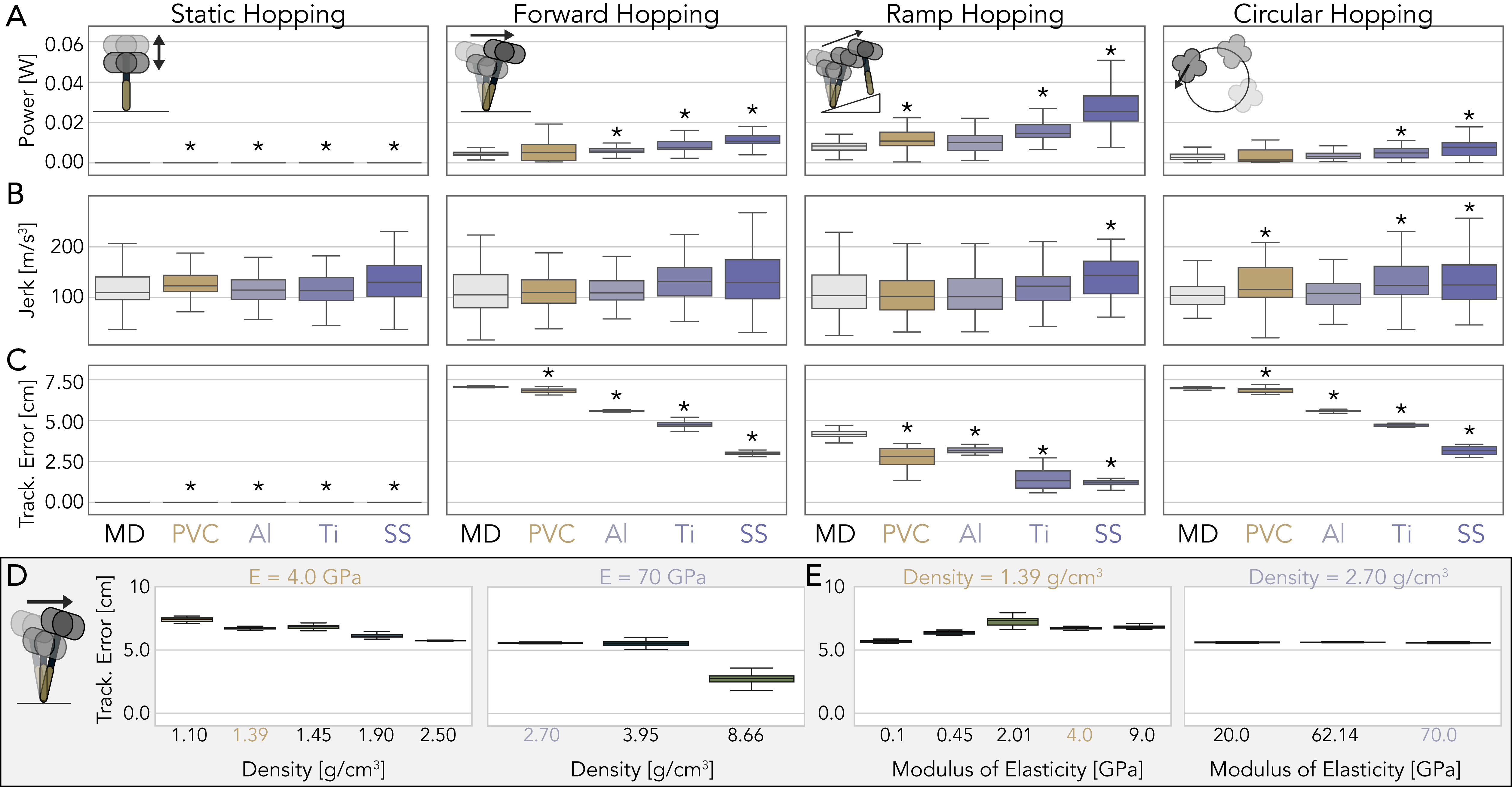}
    \caption{\textbf{Primary mono-material results}. A-C) plots illustrate the performance measures of A) power as defined in Eq. \ref{eq: power}, B) jerk as defined in Eq. \ref{eq: jerk}, and C) tracking error as defined in Eq. \ref{eq: error}, for all robotic behaviors and across five different hopper mono-material designs based on commonly used materials in robots. D-E) Plots of tracking error for experiments where one material property variable is adjusted at a time with D) displaying density shifts and E) displaying moduli shifts. Statistical significance when compared to MD (a single rigid link with $\rho = 1$ g/cm$^3$) is denoted by an asterisk ($^\star~p < 0.05)$. } \vspace{-16pt}
    \label{fig:4}
\end{figure*}
\subsection{Numerical Stability of Simulated Materials}
One challenge with modeling physical materials is that to maintain computational stability of the natural dynamics, the simulation time steps must be reduced.  Motivated by the Ashby plots, we explore the relationship between material density ($\rho$), material modulus (E), and the maximum allowable time step to maintain stable numerical computations of the simulated dynamics. This maximum required time step is obtained experimentally by iteratively reducing the simulation time step until the simulations can run without errors caused by numerical instability. The heatmap illustrated in \fig{fig:heatmap} was obtained by repeating this procedure across a grid of 400 unique density/modulus ($\rho$, E) material combinations. 

The results indicate that the time step required for stable simulations scales log-linearly with both material density and material modulus. Materials with either a low density or a high modulus require much smaller (slower) time steps than materials that are either more dense or less stiff. The intuition here is that when using fast (large) time steps with lightweight or stiff materials, the spring-like dynamics of the internal forces cause numerical instability. 

Overall, the least demanding  materials to simulate are some of the least common materials used for robotic linkages~\cite{ficht_bipedal_2021}, including soft polymers like PVC, which require a time step of 1e-3 seconds, whereas commonly used metals like Aluminum or Stainless Steel require an order of magnitude smaller time steps (1e-4 seconds) for stable simulation (\fig{fig:heatmap}). In comparison, the default time step in MuJoCo is 2e-3 seconds.

\section{Bio-inspired materials on a Robotic Hopper}
\label{sec: results}
To evaluate the hypothesis that materials influence the functional performance of robotic systems, we first apply our approach to simulate the one-legged hopper across four different materials: PVC, Aluminum (Al), Titanium (Ti), and Stainless Steel (SS). We compare the resulting performance with the standard rigid-body model. We term the standard model as \textit{MuJoCo Default} (MD) since it uses the default MuJoCo settings for density ($\rho = 1$ g/cm$^3)$. We repeat these performance comparisons across four diverse robotic behaviors: (1) static hopping, (2) forward hopping, (3) ramp hopping, and (4) circular hopping. These behaviors are illustrated in \fig{fig:methods}D and demonstrated in \fig{fig:tiles}. 

\begin{figure*}[t] 
    \centering
\includegraphics[width=0.95\textwidth]{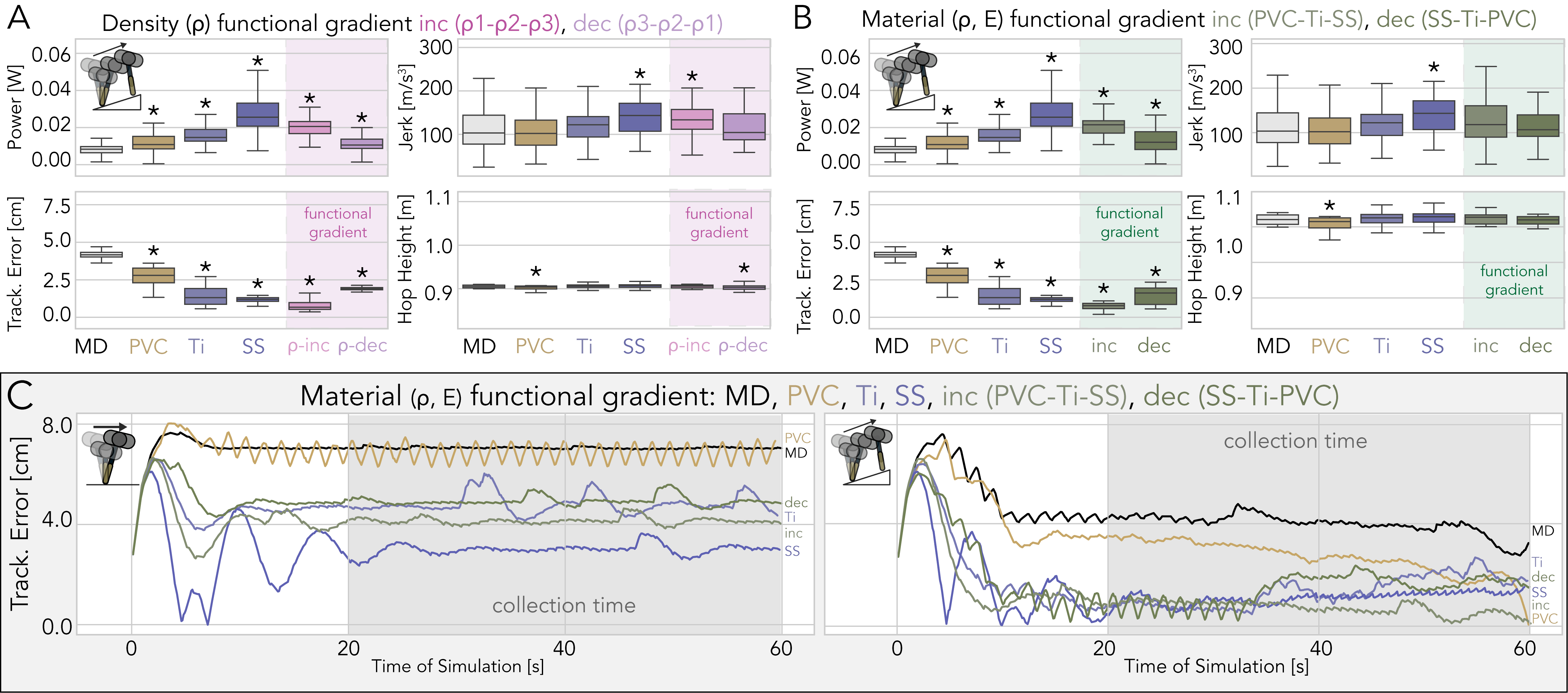}
\caption{\textbf{Functional gradient material results.} A) Functional gradient designs shown by \fig{fig:methods}A with ($\rho_1$-$\rho_2$-$\rho_3$) and ($\rho_3$-$\rho_2$-$\rho_1$) for $(\rho_1,\rho_2,\rho_3) = (1.39,4.43,8) ~$g/cm$^3$. B) Functional gradient design with varying order of materials (PVC-TI-SS and SS-Ti-PVC). C) Time-series data across all evaluated materials for forward hopping (left) and ramp hopping (right). \vspace{-17pt} }
    \label{fig:5}
\end{figure*}

To evaluate the functional performance of our robotic system, we evaluate four key metrics:
\begin{align}
    \textrm{power}_i &= \frac{1}{N_i}\sum_{t = t_0^i}^{t_f^i}\| \boldsymbol{\tau}(t) \cdot   \dot{\boldsymbol{\phi}}(t) \|_2, \label{eq: power}\\
    \textrm{tracking error}_i &= \frac{1}{N_i} \sum_{t = t_0^i}^{t_f^i}\|\mathbf{X}(t) - \mathbf{X}_d(t) \|_2, \label{eq: error}\\
    \textrm{jerk}_i &= \frac{1}{N_i}\sum_{t = t_0^i}^{t_f^i} | \dddot{p}_z(t) |, \label{eq: jerk} \\
    \textrm{hop height}_i &= h^{\textrm{act}}_{i} := \{\textrm{max}(p_z(t)) \mid t \in [t_0^i, t_f^i]\}. \label{eq: height}
\end{align}
Each metric is evaluated for an individual hop cycle $i$, indexed by the time interval $t \in [t_0^i, t_f^i] \subset \R_+$, with the first three being taken as the mean over $N_i$ total time steps in the interval. To avoid the transient behavior at the start of each simulation (illustrated by the time-series plots in \fig{fig:5}C), the metrics were computed after the initial 20 seconds, with the full duration of each simulation lasting 60 seconds. The box-and-whisker plots of \fig{fig:4}A-E and \fig{fig:5}A-B illustrate the mean and variance of the computed metrics across all hop cycles.

\subsection{Material inclusion reduces tracking error}

The results of the simulated comparisons across the four aforementioned physical materials and the rigid-body design (MD) are illustrated in \fig{fig:4}A-C. Hop height is not included in the figure since all experiments had a mean hop height of 1 m. For all behaviors, the hopper design with physical materials yields a significant decrease in tracking error, with at least a 55\% reduction for SS across all behaviors. This statistical significance is determined by using a Mann-Whitney U Test between the physical material and the MD design. However, for the more dynamic behaviors (2-4), this decrease in tracking error is afforded at the expense of significantly increased power consumption, with at most a 193\% increase in power consumption for SS across all behaviors.

\subsection{Density and Porosity shifts allow functional tune-ability}
Since the physical materials evaluated in the aforementioned experiment have two dependent variables ($\rho$ and E), we conduct a second set of experiments that only vary one property at a time, with the utilized ranges provided in (\fig{fig:4}D-E). The results of these experiments demonstrate that material density has a more noticeable influence on tracking error compared to material modulus. 

When holding the elastic modulus constant but shifting porosity, we see that even small shifts of 50\% in porosity cause large decreases in tracking error. In comparison, material modulus does not appear to significantly influence the tracking error except when considering materials across the two different classification groups, where materials with lower moduli values (E $\leq 9$ GPa i.e., polymers and natural materials) have much higher tracking error compared to stiffer materials (E $\geq 20$ GPa i.e., metals and ceramics). 

\subsection{Multi-material links provide functional compromises}
Lastly, to evaluate our hypothesis that anisotropic materials offer functional advantages compared to isotropic designs, we conduct a third set of experiments in which the leg of the hopper is constructed as an axially functional gradient link~\cite{liu_functional_2017}. Specifically, four different functional gradient designs are evaluated: 1) increasing link densities ($\rho_1$-$\rho_2$-$\rho_3$ corresponding to density values of $(1.39,~4.43,~8)~$g/cm$^3$ with $\rho_1$ located closest to the hopper base frame and $\rho_3$ located closest to the foot); 2) decreasing link densities whereby the order of the links is reversed ($\rho_3$-$\rho_2$-$\rho_1$); 3) varying materials (across both density and modulus) in order of increasing modulus (PVC-Ti-SS); and 4) varying materials in the reverse order (SS-Ti-PVC).

The results of these experiments and how they compare to the prior experimental results with isotropic materials are illustrated in \fig{fig:5}. The results with the anisotropic materials still yield significant functional advantages for tracking error but with a reduced power consumption compared to isotropic materials. For example, when considering ramp hopping, the functional gradient with increasing density yields an 82\% reduction in tracking error at only a 123\% increase in power consumption, compared to SS, which only offered a 72\% reduction in tracking error at a 193\% increase in power. In viewing material gradients (\fig{fig:5}B), we see similar benefits; functional gradients provide a reduction of power, tracking error, and jerk from SS counterparts across the entire simulation time (\fig{fig:5}C). 

Additionally, the results indicate that the direction of the functional gradients impacts the performance. Moreover, the fact that $\rho$-inc and $\rho$-dec are unique  (\fig{fig:5}A), as well as the material functional gradients (\fig{fig:5}B), implies that materials are interacting with each other, not simply combining into one bulk \textit{equivalent} material, which could be implemented in systems through multi-material 3D printing~\cite{smith_digital_2024}.

\section{Conclusions and Future Work}
\label{sec: conclusions}
In this paper, we utilize simulation as a method for evaluating how bio-inspired material relationships impact robotic behaviors across various locomotive tasks. Our results find that material properties significantly influence robotic performance, with isotropic materials yielding functional trade-offs. Our results also determine that material gradients provide several functional advantages over isotropic/homogeneous designs. Materials with increasing density gradients yield reduced tracking error, reduced power consumption, and reduced jerk compared to mono-materials. These functional gradients, while abundant in nature (e.g., tree stems, mammal skin, and the Achilles tendon), are rarely utilized in mechanical design. Our results show future designs should leverage the functional benefits of biologically inspired materials. 

Notable limitations of our methodology include the slow simulation speeds required to model porous or stiff materials. This limitation is exacerbated by the fact that very few simulation tools currently model physical material properties in computationally tractable ways. This work demonstrates the need to model physical material in robotics systems by explaining the importance of material selection on robotic performance. Future work also includes expanding the performance metrics to others such as information entropy \cite{haeufle_quantifying_2014}.


The key takeaway of this work is that the inherent complexities of materials drive functional improvements in robotic systems. This motivates further study, particularly of anisotropic materials, to understand how physical properties influence performance. It also highlights the importance of advancing research and developing simulation tools to model these complexities for real-world applications.
\vspace{-4pt}

\section*{ACKNOWLEDGMENTS}
This work was supported by the Max Planck Society and the Georgia Institute of Technology. 

\clearpage

\bibliographystyle{ieeetr}
\balance 
\bibliography{./Bibliography/IEEEabrv, ./Bibliography/references}

\end{document}